\documentclass{article}
\usepackage{spconf, amsmath, amssymb, graphicx}
\usepackage{paralist, url}
\usepackage{algpseudocode}
\usepackage{algorithm}
\usepackage{epsfig, bm}
\usepackage{xcolor}
\usepackage{hyperref}

\ninept

\title{Pykaldi2: Yet another speech toolkit based on Kaldi and Pytorch}
\name{Liang Lu,
\thanks{Code is available at: \url{https://github.com/jzlianglu/pykaldi2.git}}
Xiong Xiao, Zhuo Chen, Yifan Gong}
\address{Microsoft Speech and Language Group\\
\texttt{\small \{liang.lu, xiong.xiao, zhuc, yifan.gong\}@microsoft.com}%
}

\begin{document}
\maketitle

\begin{abstract}
We introduce PyKaldi2 speech recognition toolkit implemented based on Kaldi and PyTorch. While similar toolkits are available built on top of the two, a key feature of PyKaldi2 is sequence training with criteria such as MMI, sMBR and MPE. In particular, we implemented the sequence training module with on-the-fly lattice generation during model training in order to simplify the training pipeline. Furthermore, a preliminary version of lattice-free MMI training has also been implemented in the toolkit. To address the challenging acoustic environments in real applications, PyKaldi2 also supports on-the-fly noise and reverberation simulation to improve the model robustness. With this feature, it is possible to backpropogate the gradients from the sequence-level loss to the front-end feature extraction module, which, hopefully, can foster more research in the direction of joint front-end and backend learning. We performed benchmark experiments on Librispeech, and show that PyKaldi2 can achieve competitive recognition accuracy.  The toolkit is released under the MIT license. 
 
\end{abstract}

\section{Introduction}

In the past few years, there has been a tremendous progress in both research and applications of the speech recognition technology, which can be largely attributed to the adoption of deep learning approaches for speech processing, as well as the availability of open source speech toolkits such as Kaldi~\cite{povey2011kaldi}, PyTorch~\cite{PyTorch}, Tensorflow~\cite{tensorflow}, etc. As the most popular open-source speech recognition toolkit, Kaldi has its own deep learning library and the neural network training recipe, yet, there are persistent demands to connect Kaldi with the mainstream deep learning toolbox such TensorFlow and PyTorch. Firstly, the connection will enable the inference of Kaldi models in the environment of TensorFlow or PyTorch, which is particularly desirable from the perspective of speech applications. Secondly, the connection can give access to the rich set of APIs in TensorFlow or PyTorch for training of Kaldi models such as the distributed parallel training package. Last but not least, since the deep learning toolkits are widely used other modalities such image and natural language,  the connection may be particularly useful for multimodal learning with speech such as audio-visual processing.

In this work, we introduce PyKaldi2, which combines the strengths of Kaldi and PyTorch for speech processing. The toolkit is built on the PyKaldi~\cite{can2018pykaldi} --- the python wrapper of Kaldi. While there has been similar toolkits  built on top of Kaldi and PyTorch such as~\cite{ravanelli2019pytorch}, PyKaldi2 is different in the sense of a deeper integration of Kaldi and PyTorch, thanks to the python wrapper of Kaldi. In particular, the toolkit of~\cite{ravanelli2019pytorch} only supports model training with the cross-entropy (CE) criterion at the point of writing, while PyKaldi2 supports both CE training as well as sequence discriminative (SE) training with criteria including minimum mutual information (MMI), minimum phone error (MPE) as well as state-level minimum Bayes risk (sMBR). To simplify the sequence training pipeline, we adopt the workflow of on-the-fly lattice generation, which involves interleaved CPU-GPU computation. While generating lattices on-the-fly during model training can simplify training pipeline, the downside is that it can also slow down the training speed. To mitigate this problem, we take advantage of the distributed parallel training frameworks in Pytorch such as the Horovod library~\cite{sergeev2018horovod}, which can significantly improve the efficiency of sequence training in PyKaldi2. To improve the model robustness, this toolkit also supports on-the-fly noise and reverberation simulation.  

We performed the benchmark experiments on Librispeech public dataset~\cite{panayotov2015librispeech}. We show that PyKaldi2 can achieve competitive recognition accuracy, and the sequence training can lead to consistent improvements. The rest of the paper is organized as follows. We first outline the architecture of PyKaldi2 in section \ref{sec:arch}. In section \ref{sec:imp}, we explain the implementation details of simulation, sequence discriminative training and distributed parallel training framework in PyKaldi2. In section \ref{sec:exp}, we show our benchmark experimental results on the Librispeech corpus. Section \ref{sec:conc} concludes this paper. 

\begin{figure}[t]
\small
\centerline{\includegraphics[width=0.4\textwidth]{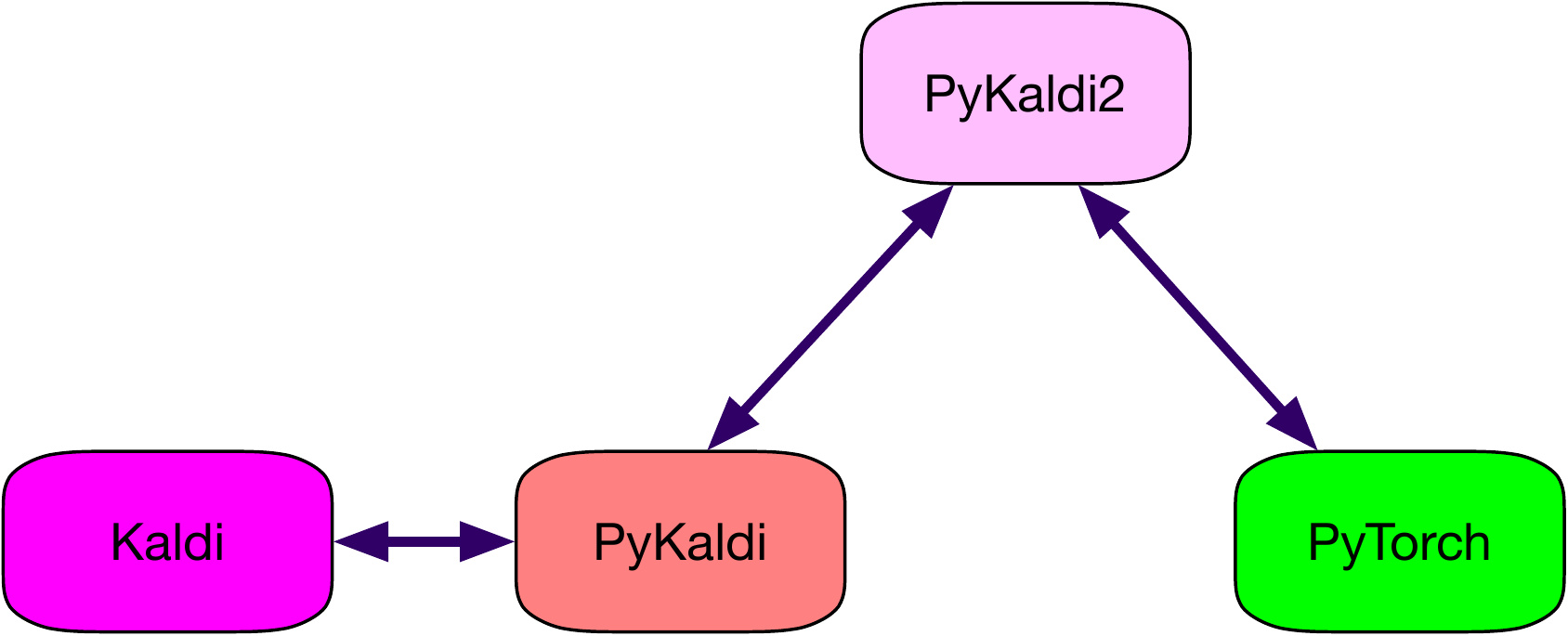}}
\caption{PyKaldi2 is built on top of Kaldi, PyKaldi and PyTorch. PyKaldi is the Python wrapper of Kaldi, which is used to access Kaldi functionalities. PyTorch is mainly used for training neural networks. }  
\label{fig:arch}
\vskip -2mm
\end{figure}

\section{Architecture}
\label{sec:arch}

\begin{figure}[t]
\small
\centerline{\includegraphics[width=0.45\textwidth]{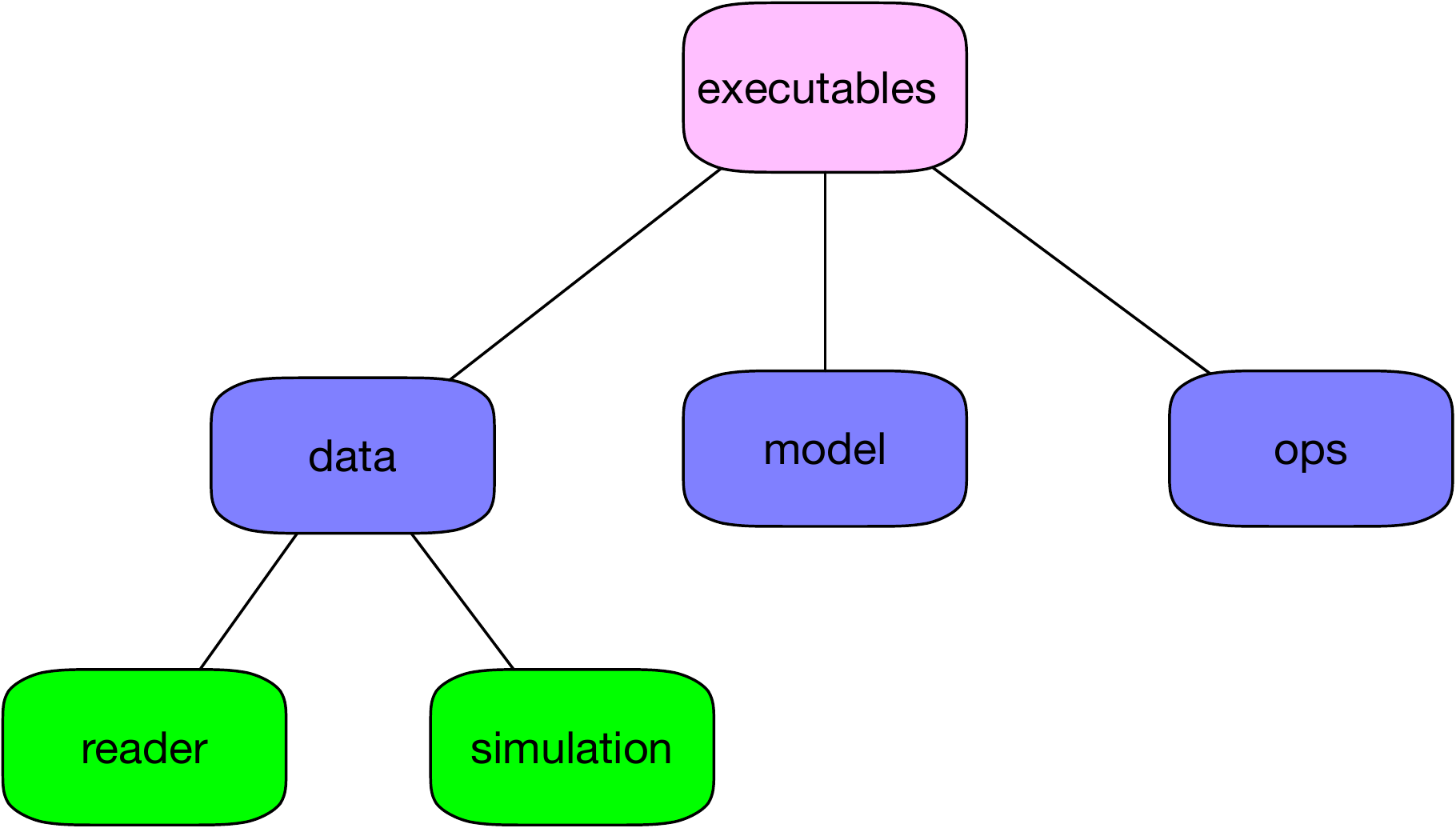}}
\caption{The layout of the modules in PyKaldi2. }  
\label{fig:layout}
\vskip -3mm
\end{figure}

The high-level architecture of PyKaldi2 is shown in Figure~\ref{fig:arch}. We take advantage of the Pykaldi to access the Kaldi functionalities. In particular, any opterations that involves the hidden Markov models (HMMs) and finite state transducers (FSTs) are performed on the Kaldi backend. PyTorch is mainly responsible for neural network training. Currently, PyKaldi2 still replies on Kaldi for bootstrapping, i.e., defining the HMM topology, building phonetic decision tree as well as generating the initial force-alignment, etc. In the current release, the key modules of PyKaldi2 are organized in the following folders, as shown in Figure~\ref{fig:layout}.

\begin{itemize}
\item {\it reader}: which implements data IO related functions. 
\item {\it simulation}: which implements the on-the-fly data simulation given the noise and room impulse response (RIR) information. 
\item {\it data}: which defines the PyTorch dataset and dataloader classes to prepare minbatches for neural network training. 
\item {\it model}: which defines the neural network acoustic models in PyTorch. 
\item {\it ops}: which implements the operations that are not available in PyTorch APIs, such as sequence training criterion operations. 
\item {\it executables}: which contains the top-level executables such CE and SE training scripts as well as lattice and alignments generation scripts. 
\end{itemize}

In the following section, we discuss the implementations details of the key modules in PyKaldi2.

\section{Implementation}
\label{sec:imp}

\subsection{On-the-fly Simulation}
\label{sec:simu}

Data simulation is an effective approach to improve the model robustness in adverse environments. A typical simulation process includes convolving the clean speech waveforms with an RIR and adding the output with different types of noise to produce reverberant and noisy speech. In static simulation, the simulated speech is generated prior to model training with a limited range of combinations of noise conditions and RIRs. In PyKaldi2, we implemented the on-the-fly data simulation, which randomly selects a noise condition and an RIR for each utterance during model training. The advantage is that for different epochs, the model can see data from different combinations of noise and reverberation conditions, which can improve the model robustness in a more efficient and flexible manner. PyKaldi2 supports both single channel and multi-channel data simulation, and both single speaker source and multiple speaker source simulation. 

\subsection{On-the-fly Lattice Generation}  
\label{sec:lat}

In a typical sequence training pipeline such as in Kaldi, the lattices are usually generated in a batch fashion first using a seed (CE) model, sequence training is then performed by computing losses by rescoring the lattices. The sequence training pipeline in PyKaldi2 is different in the sense that the lattices are always generated on-the-fly during model training. One reason of this design is to make the sequence training pipeline simpler. More importantly, the approach may work better for very large scale dataset, as the lattices are always produced from the well-matched model. The skeleton of the sequence training pipeline is shown in Table \ref{tab:se-train}. 

\subsection{On-the-fly Alignment}
\label{sec:ali}

Acoustic model training usually involves refreshing the force alignments multiple times when better acoustic models or more training data are available such as the standard training pipeline in Kaldi. While this pipeline also works in PyKaldi2, we also implemented the model training with on-the-fly alignment generation. With the similar motivation for online lattice generation, this recipe can reduce the complexity for system building, and may even improve the quality of the model in certain cases with a large amount of training data. An example of sequence training with both on-the-fly lattices and alignments is shown in Table~\ref{tab:se-train2}.

\begin{table}[t]\centering
\caption{Sequence training pipeline in Pykaldi2 with static alignments.}
\label{tab:se-train}
\footnotesize
\vskip0.15cm
\begin{tabular}{l}
\hline 

\hline

Input:  \\
\hskip4mm {\tt model} - a seed neural network acoustic model \\
 \hskip4mm $X_{1:T}$ - acoustic frames indexed from 1 to T\\
 \hskip4mm {\tt log\_prior} - log prior for each tied HMM state\\
 \hskip4mm {\tt alignment}  -  force alignment in the form of transition indexes\\
  \hskip4mm {\tt HCLG} -  Kaldi decoding graph for lattice generation \\
 \hskip4mm {\tt word\_ids}: word to index mapping \\
 \hskip4mm {\tt trans\_model}: Kaldi transition model \\
  \\ 
1: {\it Setup Kaldi decoder} \\
\\[-1em]
 {\hskip4mm\color{gray} \# call Pykaldi API} \\
 \hskip4mm asr\_decoder = MappedLatticeFastRecognizer ({\tt trans\_model}, \\ 
 \hskip5.5cm  {\tt HCLG},  \\
 \hskip5.5cm {\tt word\_ids}, \\
 \hskip5.5cm {\tt decoder\_opts})\\
 
 2: {\it Compute log-likelihoods  } \\
 \\ [-1em]
 {\hskip4mm \color{gray} \# running on GPU with PyTorch } \\
 \hskip4mm log\_likes = {\tt model}(X) - {\tt log\_prior} \\
 \\
 
 3: {\it Lattice generation} \\
 {\hskip4mm \color{gray} \# call PyKaldi API, running on CPU} \\
 \hskip4mm decoder\_out = asr\_decoder.decode(log\_likes)  \\ 
 \\ 
 4: {\it Compute the gradients for backpro } \\
 \\[-1em]
 {\hskip4mm \color{gray} \# call PyKaldi API, running on CPU} \\
 \hskip4mm loss = LatticeForwardBackward ({\tt trans\_model},  \\
 \hskip4.15cm log\_likes, \\
 \hskip4.15cm decoder\_out[``lattice"], \\
 \hskip4.15cm {\tt alignment}, \\
 \hskip4.15cm  criterion=``mmi $\mid$ smbr $\mid$ mpe")  \\
 \\[-1em]
 \hskip4mm loss.cuda() {\hskip4mm \color{gray} \# move the gradients to GPU} \\
 \\
 5: {\it Model update} \\
 \\[-1em]
 \hskip4mm loss.backward() {\hskip4mm \color{gray} \# running on GPU with PyTorch} \\
 \hskip4mm optimizer.step() \\
\hline

\hline
\end{tabular}
\vskip-4mm
\end{table}

\subsection{Distributed Parallel Training}
\label{sec:distributed}

The sequence training in PyKaldi2 involves cross CPU-GPU computation as shown in Table~\ref{tab:se-train}. The training speed can be low if we only use a single thread due to online lattice generation. To mitigate this problem, we use the Horovod library -- an implementation of the more efficient ring-allreduce synchronization technique -- for distributed parallel training. In our benchmark experiments, it can significantly improve the training speed by leveraging on multiple GPUs and CPUs. In the future, we shall also implement the lattice generation with multiple CPU threads, which is currently the bottleneck hindering the training speed.

\begin{table}[t]\centering
\caption{Sequence training pipeline in Pykaldi2 with on-the-fly alignment generation.}
\label{tab:se-train2}
\footnotesize
\vskip0.15cm
\begin{tabular}{l}
\hline 

\hline

Input:  \\
\hskip4mm {\tt model} - a seed neural network acoustic model \\
 \hskip4mm $X_{1:T}$ - acoustic frames indexed from 1 to T\\
 \hskip4mm {\tt log\_prior} - log prior for each tied HMM state\\
 \hskip4mm {\tt text}  -  word-level ground truth transcription\\
 \hskip4mm {\tt tree} - desicion tree for HMM state tying \\
 \hskip4mm {\tt L.fst} - lexicon fst \\
  \hskip4mm {\tt HCLG} -  Kaldi decoding graph for lattice generation \\
 \hskip4mm {\tt word\_ids}: word to index mapping \\
 \hskip4mm {\tt trans\_model}: Kaldi transition model \\
  \\ 
1: {\it Setup Kaldi decoder and aligner} \\
\\[-1em]
 {\hskip4mm\color{gray} \# call Pykaldi API} \\
 \hskip4mm asr\_decoder = MappedLatticeFastRecognizer ({\tt trans\_model}, \\ 
 \hskip5.5cm  {\tt HCLG},  \\
 \hskip5.5cm {\tt word\_ids}, \\
 \hskip5.5cm {\tt decoder\_opts})\\
 
 \hskip4mm aligner = MappedAligner.from\_files({\tt trans\_model}, \\
 \hskip4.5cm {\tt tree}, \\
 \hskip4.5cm  {\tt L},  \\
 \hskip4.5cm {\tt align\_opts})\\
 
 2: {\it Compute log-likelihoods  } \\
 \\ [-1em]
 {\hskip4mm \color{gray} \# running on GPU with PyTorch } \\
 \hskip4mm log\_likes = {\tt model}(X) - {\tt log\_prior} \\
 \\
 
 3: {\it Lattice generation} \\
 {\hskip4mm \color{gray} \# call PyKaldi API, running on CPU} \\
 \hskip4mm decoder\_out = asr\_decoder.decode(log\_likes)  \\ 
 \\ 
 
 4: {\it Alignment generation} \\
 {\hskip4mm \color{gray} \# call PyKaldi API, running on CPU} \\
 \hskip4mm align\_out = aligner.align(log\_likes, {\tt text})  \\ 
 \\ 
 
 5: {\it Compute the gradients for backpro } \\
 \\[-1em]
 {\hskip4mm \color{gray} \# call PyKaldi API, running on CPU} \\
 \hskip4mm loss = LatticeForwardBackward ({\tt trans\_model},  \\
 \hskip4.15cm log\_likes, \\
 \hskip4.15cm decoder\_out[``lattice"], \\
 \hskip4.15cm align\_out[``alignment"], \\
 \hskip4.15cm  criterion=``mmi $\mid$ smbr $\mid$ mpe")  \\
 \\[-1em]
 \hskip4mm loss.cuda() {\hskip4mm \color{gray} \# move the gradients to GPU} \\
 \\
 5: {\it Model update} \\
 \\[-1em]
 \hskip4mm loss.backward() {\hskip4mm \color{gray} \# running on GPU with PyTorch} \\
 \hskip4mm optimizer.step() \\
\hline

\hline
\end{tabular}
\vskip-4mm
\end{table}

\section{Experiments}
\label{sec:exp}

We performed the benchmark experiments using the publicly available Librispeech corpus~\cite{panayotov2015librispeech}, which contains around 960 hours of training data in total. Most of our experiments were performed on a single machine with 4 Tesla V100 GPUs unless specified otherwise. In terms of the neural network model, we used a standard bidirectional LSTM for acoustic modeling, which has 3 hidden layers, and each layer has 512 hidden units. The total number of parameters is around 21 million. We presented some additional results from a transformer-based acoustic model with on-the-fly lattice generation for sequence training in~\cite{lu2019transformer}. We used 80-dimensional raw log-mel filter-bank features, and we did not do any form of speaker-level feature normalization. Instead, we only applied the utterance-level mean and variance normalization. We used a 4-gram language model for decoding that is released as the part of the corpus. We used Kaldi to build a Gaussian mixture model (GMM) system for bootstrapping.   

\begin{table}[t]\centering
\caption{Results from models trained with 960 hours of Librispeech data.}
\label{tab:960hr}
\footnotesize
\vskip0.15cm
\begin{tabular}{l|ccccc}
\hline 

\hline
model       &  loss     &  dev-clean       &  dev-other & test-clean & test-other \\ \hline
  & CE & 4.6 & 13.3 & 5.1 & 13.5  \\
BLSTM & MMI  & 4.3   & 12.1  & 4.8 & 12.5 \\
  & sMBR & 4.3 & 12.3 & 4.9 & 12.5 \\
  & MPE  & 4.4 & 12.3 & 4.8 & 12.5  \\

\hline

\hline
\end{tabular}
\vskip-4mm
\end{table}

\begin{table}[t]\centering
\caption{Static versus dynamic alignment for sequence training. The model was trained with 960 hours of training data and MMI criterion, and it was tested on {\tt dev-other} evaluation set.}
\label{tab:ali}
\footnotesize
\vskip0.15cm
\begin{tabular}{l|ccccc}
\hline 

\hline
  & \multicolumn{5}{c}{\#steps}  \\
  alignment & 0 & 1000 & 2000 & 4000 & 8000 \\ \hline
   static & 13.32 & 12.28 & 12.24 & 12.17 & 12.14  \\
  dynamic & 13.32  & 12.33   & 12.31  & 12.28 & 12.24 \\
\hline

\hline
\end{tabular}
\vskip-4mm
\end{table}

\subsection{Benchmark results}
\label{sec:960}

We report benchmark results of PyKaldi2 using 960 hours of clean training data from Librispeech. The phonetic decision tree and force-alignments were obtained from a Kaldi GMM system. The total number of tied HMM states is 5768. For CE training, we used the Adam optimizer with an initial learning rate as $2\times 10^{-4}$, and decayed the learning rate by a factor of 0.5 from the $4^{th}$ epoch. We trained the model for 8 epochs. We then used the CE model as the seed, and fixed the learning rate as $1 \times 10^{-6}$ and used the vanilla SGD optimizer for MMI, sMBR and MPE training. For sequence training, we used 4 GPUs with Horovod and each GPU consumed 4 utterances for each update, corresponding to the minibatch size as 16 utterances. To avoid overfitting, we applied CE regularization with weight as $0.1$. We first run experiment with fixed alignments from the GMM system. In our experiments, we observed the sequence training converged very quickly, and the optimal results were achieved at around 8000 steps. The word error rate (WER) results are given in Table~\ref{tab:960hr}. Overall, sequence training can achieve around 5-6\% relative WER reduction in the clean condition, and over 7\% in the {\tt other} condition, which demonstrates that the sequence training pipeline of PyKaldi2 works as expected. 

We then evaluated the sequence training pipeline with dynamic alignment in PyKaldi2 as shown in Table \ref{tab:se-train2}. In our Librispeech experiments, we observed that the model trained with approach can overfit very quickly. We have to increase the CE regularization from 0.1 to 0.4 and apply the L2 regularization with the weight as $0.001$ to mitigate the overfitting problem. In Table \ref{tab:ali}, we compare the sequence training results on {\tt dev-other} evaluation sets with model trained with and without dynamic alignments using the MMI criterion. Although the convergence trends of the two approaches are similar, the dynamic alignments approach did not result in gains in terms of accuracy, contrary to our expectation. In the future, we shall investigate if the overfitting behavior of dynamic alignment depends on the acoustic model and the dataset.

\subsection{Training speed}

We then show some information regarding the training speed of PyKaldi2. While the training speed is highly variable depending on factors including the type of GPUs that are used to train the model, the type and the size of the acoustic model, data connection between GPUs, as well as the size of each minibatch etc, we show some statistics based on our experimental configuration to give the readers a glimpse of the training speed of PyKaldi2. To evaluate the training speed, we use the metric of the inverted real time factor (iRFT) in training, which measures how many hours of data we can process per hour. Note that, the acoustic sampling rate in our experiments is 100 Hz. Given that, we can also convert the iRTF to the metric of the number of frames per second. Lower sampling rates usually improve the training speed significantly. Table~\ref{tab:speed} shows the speed of both CE and MMI training. The iRTFs of sMBR and MPE training are very similar to those of MMI. For CE training, the sequence length of each minibatch is the size of the context window, which is set to be 80 in our experiments. For SE training, the sequence length of each minibatch is variable, which is the length of the longest utterance in the minibatch.

\begin{table}[t]\centering
\caption{Statistics of the training speed with different number of GPUs and minibatch size.}
\label{tab:speed}
\footnotesize
\vskip0.15cm
\begin{tabular}{l|cccc}
\hline 

\hline
model       &  loss     &  batch size        &  \#GPUs  & iRTF  \\ \hline
  & CE & 64 & 1 & 190  \\
  & CE & 64 & 4 & 220  \\
  & CE & 256 & 4 & 520 \\
  & CE & 1024 & 16 & 1356 \\
BLSTM & MMI	  & 1   & 1  & 11.6     \\
 & MMI	  & 2   & 1  &  13.4    \\
 & MMI	  & 4   & 1  &  16.7    \\
 & MMI	  & 4   & 4  & 34.5      \\
 & MMI	  & 8   & 4  & 44.2      \\
 & MMI	  & 16   & 4  & 50.4      \\
 & MMI    & 64  & 16 & 176 \\

\hline

\hline
\end{tabular}
\vskip-4mm
\end{table}

Table \ref{tab:speed} shows the training speed for both single GPU and multi-GPU jobs. For both CE and SE training, the speedup from multi-GPU training is substantial. When training with only 1 GPU, increasing the batch size from 1 to 4 only resulted in around 1.5 times speedup for SE training. The reason is that the lattice generation and forward-backward computation are not parallelized in each minibatch in the current implementation. With multi-GPU training by Horovod, we can run lattice generation as well as the computation of the SE loss in parallel, which dramatically improves the training speed. For example, if we set the minibatch size for each GPU to be 4, we can process 16 utterances with 4 GPUs for each model update, and the training speed can reach to around 50 hours of training data per hour. With 16 GPUs, the iRTF can reach to over 170.

\subsection{Results from on-the-fly simulation}

To evaluate the simulation module of PyKaldi2, we collected a dataset\footnote{The dataset will be made publicly available in the near future.} by replaying the Librispeech {\tt dev-clean} evaluation data in a meeting room. The dataset consists of simulated meetings from multiples speaker that are replayed by loud speakers. The data was then recorded by a microphone array in the middle of the meeting room. The dataset has 10 hours of recording in total, and it is split into 10 sessions. In each session, there are 6 mini-meetings, and each mini-meeting has 10 minutes of recording with overlap ratios ranging from 0 to 40\%. For our evaluation, we only tested the model on the data recorded from the central microphone, which a typical far-field condition without beamforming.  

In our data simulation experiment, we used noise from various sources and simulated RIR. In particular, the additive noise files from CHiME-4 challenge \cite{vincent2017analysis}, DEMAND noise database \cite{thiemann2013demand}, MUSAN corpus \cite{snyder2015musan}, Noisex92 corpus \cite{noisex92}, DCASE 2017 \cite{mesaros2017dcase}, and the Non-speech OSU corpus \cite{hu2010tandem}. The RIRs are all simulated, each with sampled room size, source and microphone positions, and T60 reverberation time. The T60 time is sampled uniformly from 0.2 to 0.5s. Table~\ref{tab:simu} shows the results of the models trained with and without data simulation. In the far-field condition, the CE-trained model with 460 hours of clean speech with simulation performed even better the MMI-trained model with 960 hours of data, which demonstrates that data simulation module in PyKaldi2 works reasonably well. In this experiment, we performed the dynamic simulation for every minibatch, and consequently, our model did not perform well in the close-talk condition due to mismatch. In the toolkit, we have introduced an option to control the simulation frequency so that we can adjust the percentage of the raw acoustic data for every training epoch. However, we did not retrain the model by tuning the simulation frequency as the focus of the paper is not to achieve the best results on this simulated evaluation set.

 \begin{table}[t]\centering
\caption{Results from the model trained with data simulation.}
\label{tab:simu}
\footnotesize
\vskip0.15cm
\begin{tabular}{ccccc}
\hline 

\hline
data & loss   & simulation     &  condition     &  WER         \\ \hline
460hr & MMI & $\times$ & FF & 17.6   \\
960hr & MMI & $\times$ & FF  & 11.8      \\
460hr & CE & $\surd$ & FF  & 11.6    \\ \hline
460hr & MMI & $\times$ & CT & 5.6   \\
960hr & MMI & $\times$ & CT  & 4.9      \\
460hr & CE & $\surd$ & CT  & 7.5    \\

\hline

\hline
\end{tabular}
\vskip-4mm
\end{table}

\subsection{Lattice-free MMI training}
\label{ssec:lfmmi}

The core functionalities for lattice-free MMI (LFMMI)~\cite{povey2016purely} training have also been implemented within the toolkit. The forward-backward algorithm in LFMMI runs on GPU in the implementation of Kaldi~\cite{povey2011kaldi}. To fully leverage on the computational power of GPUs, the training recipe in Kaldi segments the utterances in the training data into chunks of the same length, e.g., 1.5 seconds, and each minibatch has multiple sequences, e.g. 128. To reproduce similar type of training recipe in PyKaldi2, it requires a new type of dataloader which has not been done yet. We tested the LFMMI module in PyKaldi2 by using the same dataloader used for lattice-based sequence training with minibatch size as 1, and achieved over 8\% WER on the {\tt dev-clean}, which is much worse than the lattice-based sequence training results. A new dataloader and data preprocessing recipe specifically for LFMMI will be released when they are available. 

\section{Conclusion}
\label{sec:conc}

We have introduced PyKaldi2 -- a speech toolkit that is developed based on Kaldi and PyTorch. Thanks to the python wrapper of Kaldi -- PyKaldi, PyKaldi2 enjoys the deep integration of Kaldi and PyTorch, and consequently it can support the most commonly used sequence discriminative training criteria such as MMI, sMBR and MPE. For simplicity and flexibility of the training pipeline as well as the model robustness, PyKaldi2 features on-the-fly lattice generation, alignment generation and data simulation. From the benchmark experiments on Librispeech, we have showed that the sequence training pipeline can deliever the expected gains over the CE model. We also demonstrated the effectiveness of the data simulation module in PyKaldi2 by evaluating on a replayed Librispeech evaluation data. Currently, PyKaldi2 is optimized for lattice-based sequence training. One of the future works is to improve the recipe for the lattice-free technique.  

\bibliographystyle{IEEEtran}
\bibliography{bibtex}

\end{document}